\documentclass[lettersize,journal]{IEEEtran}
\usepackage{amsmath,amsfonts}
\usepackage{algorithmic}
\usepackage{algorithm}
\usepackage{array}
\usepackage[caption=false,font=normalsize,labelfont=sf,textfont=sf]{subfig}
\usepackage{textcomp}
\usepackage{stfloats}
\usepackage{url}
\usepackage{verbatim}
\usepackage{graphicx}
\usepackage{multirow}

\usepackage{color}

\usepackage{adjustbox}

\usepackage{booktabs} 

\usepackage[table]{xcolor}
\usepackage{colortbl}

\usepackage{biblatex}  
\usepackage[colorlinks=true,linkcolor=black,citecolor=black, urlcolor=black]{hyperref}

\begin{document}

\title{Structure-Guided Allocation of 2D Gaussians for \\ Image Representation and Compression}

\author{
Huanxiong Liang, Yunuo Chen, Yicheng Pan, Sixian Wang, Jincheng Dai, Guo Lu, and Wenjun Zhang
\thanks{
Huanxiong Liang, Yunuo chen, Yicheng Pan, Sixian Wang, Guo Lu, and Wenjun Zhang are with the Department of Electronic Engineering, Shanghai Jiao Tong University, Shanghai 200240, China.

Jincheng Dai is with the Key Laboratory of Universal Wireless Communications, Ministry of Education, Beijing University of Posts and Telecommunications, Beijing 100876, China.





}
}


\IEEEpubid{
}

\maketitle

\begin{abstract}
Recent advances in 2D Gaussian Splatting (2DGS) have demonstrated its potential as a compact image representation with millisecond-level decoding. However, existing 2DGS-based pipelines allocate representation capacity and parameter precision largely oblivious to image structure, limiting their rate–distortion (RD) efficiency at low bitrates. To address this, we propose a structure-guided allocation principle for 2DGS, which explicitly couples image structure with both representation capacity and quantization precision, while preserving native decoding speed. First, we introduce a structure-guided initialization that assigns 2D Gaussians according to spatial structural priors inherent in natural images, yielding a localized and semantically meaningful distribution. Second, during quantization-aware fine-tuning, we propose adaptive bitwidth quantization of covariance parameters, which grants higher precision to small-scale Gaussians in complex regions and lower precision elsewhere, enabling RD-aware optimization, thereby reducing redundancy without degrading edge quality. Third, we impose a geometry-consistent regularization that aligns Gaussian orientations with local gradient directions to better preserve structural details. Extensive experiments demonstrate that our approach substantially improves both the representational power and the RD performance of 2DGS while maintaining 1,000+ FPS decoding. Compared with the baseline GSImage, we reduce BD-rate by 43.44\% on Kodak and 29.91\% on DIV2K.
\end{abstract}


\section{Introduction}
\IEEEPARstart{I}{mage} representation has long been a fundamental field in computer vision and signal processing. Traditional approaches based on handcrafted transforms, such as DCT~\cite{DCT}, have been widely used in image codecs like JPEG~\cite{JPEG} and JPEG2000~\cite{JPEG2000}. Although efficient and standardized, their fixed transforms and manually designed components limit adaptability to complex image content. Furthermore, in emerging application areas that require ultra-fast decoding, such as virtual and augmented reality, modern AAA games, and edge computing, these conventional methods struggle to maintain both high fidelity and extremely low latency.

With advances in neural networks, implicit neural representations (INRs) have become a powerful paradigm for learning continuous and compact representations. Recent works~\cite{COIN,NeRF,HNR} use multilayer perceptrons (MLPs) to map spatial coordinates to high-dimensional features, thereby achieving strong performance in different tasks. However, their reliance on large MLPs results in high memory usage, long training time, and slow inference. Although grid-based extensions~\cite{Neurbf,Instant-NGP} mitigate these issues through multi-resolution feature grids, these methods still face challenges in resource-constrained environments and in achieving millisecond-level decoding latency.

\begin{figure}[!t]
\centering
\includegraphics[width=3.0in]{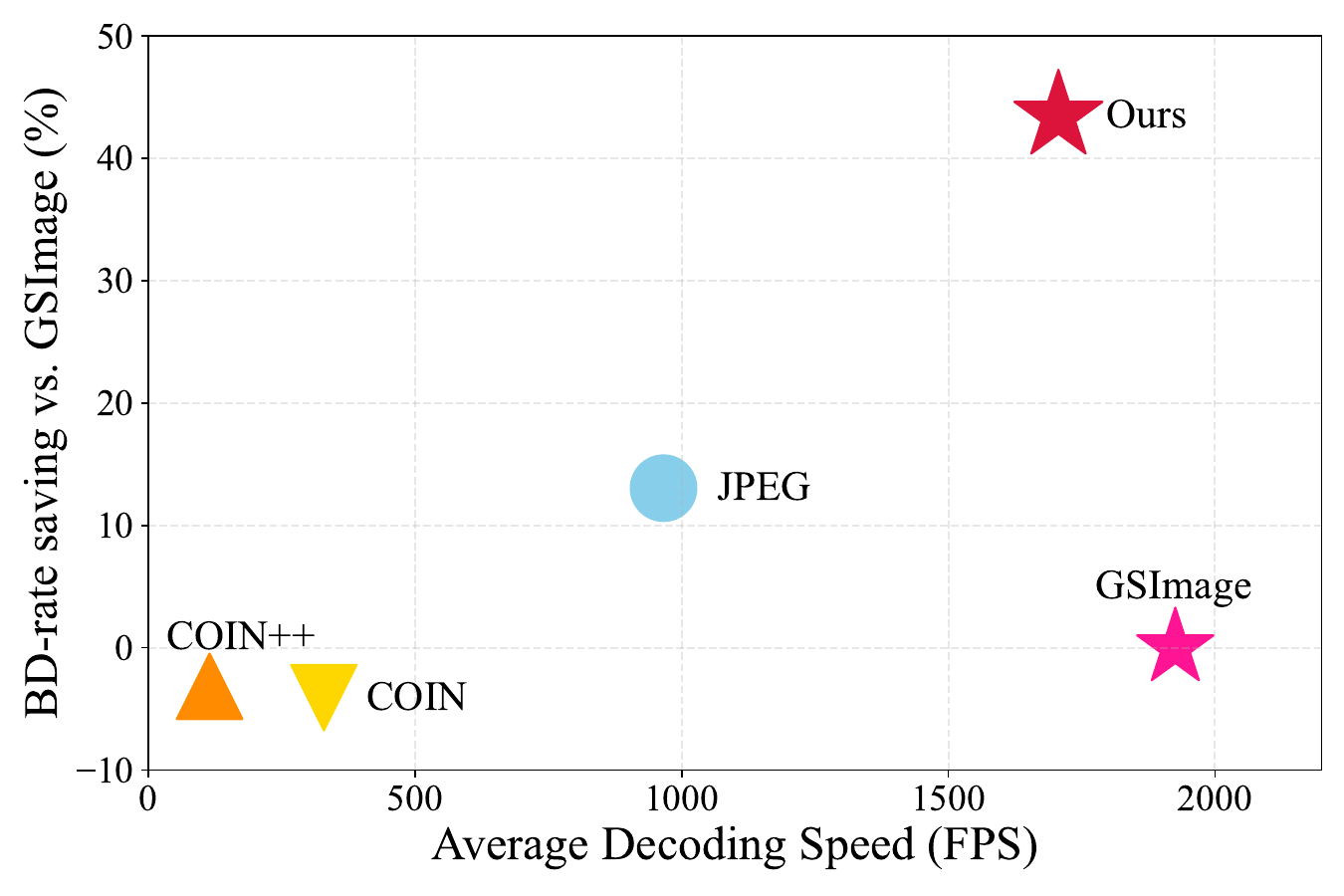}
\caption{BD-rate saving vs. Decoding Speed on Kodak dataset. Our method achieves the best BD-rate at low bitrates with similar decoding speed to GSImage. The upper-right region indicates better performance.}
\label{Compression_Comparison}
\end{figure}

Inspired by 3D Gaussian Splatting (3DGS)~\cite{3DGS},  which is known for its ultra-fast rendering speed, 2DGS~\cite{2DGS} and GSImage~\cite{Gaussianimage} extend this paradigm to 2D image representation. 2DGS introduces 2D Gaussians and applies K-means-based vector quantization for low-bitrate compression. GSImage further removes depth information, replaces $\alpha$-blending with accumulated blending, and integrates learned scale quantization with residual vector quantization for compression. These designs significantly accelerate training and decoding while achieving compression performance comparable to that of INR-based methods such as COIN~\cite{COIN} and COIN++~\cite{COIN++}.

\IEEEpubidadjcol

However, existing 2DGS pipelines allocate representational capacity and bitwidth precision in a largely structure-agnostic manner, which limits their RD efficiency at low bitrates. For example, GSImage randomly initializes 2D Gaussians without considering the spatial structural priors inherent in natural images, and applies uniform bitwidth quantization to all attributes, neglecting the varying importance of Gaussian primitives with different scales and curvatures. As a result, both representation capacity and parameter precision are inefficiently allocated across spatial regions, leading to suboptimal compression performance in detail-rich areas.

In this work, we argue that efficient compression of 2D Gaussian representations fundamentally requires a structure-guided allocation principle, which couples image structure with both representational capacity and parameter precision. First, we propose a structure-guided initialization that allocates more 2D Gaussians to structurally complex regions and fewer to smoother areas based on the variance of image gradients within each segmented region. This initialization yields a more efficient and semantically meaningful distribution of 2D Gaussians, aligning the placement of 2D Gaussians with the local visual content. During quantization-aware fine-tuning, we apply adaptive bitwidth quantization to the covariance parameters, allocating higher bitwidth precision to small-scale Gaussians in complex regions and lower precision elsewhere. By balancing precision and compactness according to spatial scale, this strategy enables RD-aware optimization, reducing parameter redundancy while preserving accurate reconstruction in detail-rich regions. Finally, we impose a geometry-consistent regularization that aligns Gaussian orientations with local gradient directions to better preserve structural details. By explicitly incorporating image structure into both representation capacity and quantization precision, our method shifts 2DGS-based image compression from uniform parameter allocation to a structure-aware RD optimization paradigm. Experimental results demonstrate that our method achieves superior representational fidelity and rate–distortion performance compared with the baseline GSImage, while maintaining a real-time decoding speed exceeding 1,000 FPS. 

Overall, our contributions are summarized as follows:

\begin{itemize}
    \item We propose a structure-guided allocation principle for 2DGS, which assigns representational capacity and parameter precision according to image structure to achieve higher fidelity and more efficient image compression.
    
    \item We realize the structure-guided allocation principle through three key designs: a structure-guided initialization that allocates Gaussians according to spatial complexity, an adaptive bitwidth quantization of covariance parameters that balances precision and compactness for RD-aware optimization, and a geometry-consistent regularization that preserves fine structural details.
    
    \item Extensive experiments demonstrate that our method substantially enhances the representational capacity of 2DGS and achieves superior rate–distortion performance compared with GSImage, while maintaining real-time decoding speeds exceeding 1,000 FPS.
\end{itemize}

\section{Related Work}
\subsection{Implicit Neural Representation}
With advances in deep neural networks, implicit neural representations (INRs) have emerged as a powerful paradigm for learning compact and continuous representations across various domains. A growing body of work~\cite{COIN, NeRF, HNR} employs MLPs as implicit functions that map spatial coordinates to data values in high-dimensional space, achieving strong performance in tasks such as image compression~\cite{HNR} and 3D reconstruction~\cite{NeRF} . However, large MLPs suffer from long training times, high memory usage, and slow inference. To mitigate these issues, subsequent grid-based methods~\cite{Tensorf,Instant-NGP} improve efficiency via multi-resolution structures but remain limited when deployed on resource-constrained devices. To address these challenges, GSImage adopts the 2DGS representation, which offers significantly faster inference and lower memory consumption. However, its ability to capture fine-grained structures remains limited, which motivates us to fully exploit the potential of 2DGS representation.

\subsection{3D Gaussian Splatting}
3D Gaussian Splatting (3DGS)~\cite{3DGS} has recently emerged as a promising technique for next-generation 3D scene reconstruction. It represents scenes with millions of learnable 3D Gaussians characterized by  geometric and appearance attributes, employing a fully differentiable, parallelizable rasterization pipeline for efficient and high-quality rendering. Unlike NeRF-based methods, 3DGS enables real-time rendering and significantly faster training, while still achieving photorealistic novel-view synthesis. Its flexibility and efficiency have led to strong performance across a wide range of applications, including high-quality 3D human generation~\cite{Humangaussian} , large-scale scene reconstruction~\cite{citygaussian}, and dynamic 3D scene reconstruction~\cite{4DGS} . GSImage extends 3DGS to 2D image representation, but it lacks image-specific initialization strategies, which limits fidelity. To address this, we propose a structure-guided initialization that allocates more Gaussians to complex regions and a geometry-consistent regularization that improves detail reconstruction, fully exploiting the potential of 2D Gaussian representations for efficient high-quality image modeling.

\subsection{Image Compression}
Conventional image codecs such as JPEG~\cite{JPEG} and JPEG2000~\cite{JPEG2000} rely on fixed transform–quantization–entropy pipelines, offering efficiency but limited adaptability to diverse image characteristics. Learned image compression (LIC) methods~\cite{COIN,COIN++} have emerged as powerful alternatives by leveraging neural networks to enable end-to-end optimization~\cite{endtoend} of the entire compression pipeline.  Most LIC approaches, which are based on VAE architectures, incorporate complex nonlinear transformations~\cite{Hyperior} and advanced entropy models~\cite{Checkerboard}, achieving superior rate-distortion performance compared with traditional image codecs. However, their complex architectures incur high computational costs and slow decoding. Recent INR-based codecs~\cite{BayesianINRcompression_guo2023} reduce model size but still struggle with scalability at high resolutions. In contrast, our method combines the strengths of INRs and LIC, achieving both strong rate-distortion performance and extremely fast decoding speed. \iftrue Building on GSImage, we introduce adaptive bitwidth quantization of the covariance parameters to enhance RD efficiency while maintaining ultra-fast, resolution-independent decoding that exceeds 1,000 FPS. \fi


\begin{figure*}[!t]
\centering
\includegraphics[width=7.0in]{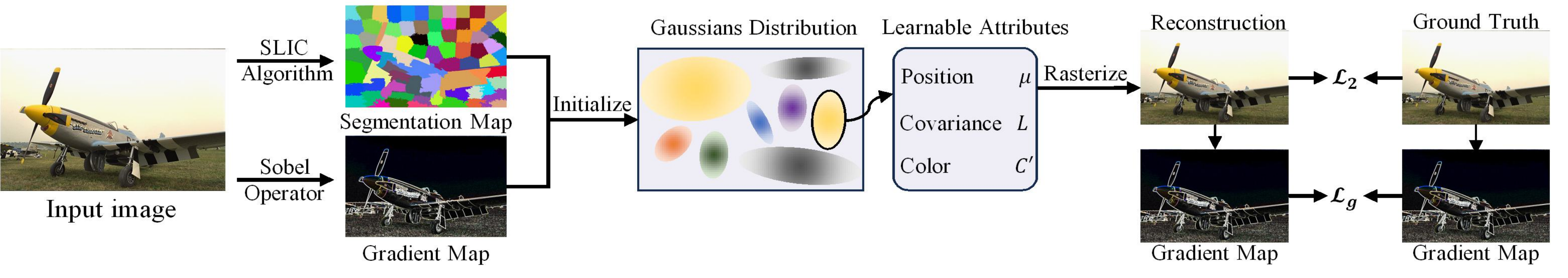}
\caption{Representation pipeline of our proposed method. We begin with structure-guided initialization, which initialize 2D Gaussians based on local structural complexity derived from the segmentation map and gradient map. These Gaussians are rasterized to reconstruct the image, and their attributes are optimized using a joint loss combining MSE loss $\mathcal{L}_2$ with geometry-consistent regularization $\mathcal{L}_g$.}
\label{Representation_Pipeline}
\end{figure*}
 
\section{Methodology}
Fig.~\ref{Representation_Pipeline} and \ref{Compression_Pipeline} present the representation and compression pipelines of our method, respectively. The representation process begins with a structure-guided initialization that allocates 2D Gaussians based on the structural complexity of the input image. These 2D Gaussians are then rasterized via accumulated blending, and their parameters are optimized using a joint loss combining MSE loss with geometry-consistent regularization. For compression, adaptive bitwidth quantization is applied to the covariance parameters during quantization-aware fine-tuning. The following subsections detail each component: Section~\ref{2DGS} reviews the 2DGS framework; Section~\ref{SGI} presents the structure-guided initialization; Section~\ref{ABQ} describes the adaptive bitwidth quantization; Section~\ref{GCR} introduces the geometry-consistent regularization; and Section~\ref{Optimization} outlines the overall training and fine-tuning procedures.

\subsection{2D Gaussian Splatting (2DGS)}\label{2DGS}
Inspired by 3DGS~\cite{3DGS}, GSImage introduces 2D Gaussian Splatting for 2D images. Each 2D Gaussian is parameterized by a position vector $\mu \in \mathbb{R}^2$, a covariance matrix $\Sigma \in \mathbb{R}^{2 \times 2}$, and color coefficients $c' \in \mathbb{R}^3$. To guarantee the positive semi-definiteness of $\Sigma$ during training, GSImage employs the Cholesky decomposition:
\begin{equation}\label{Cholesky_Factorization}
    \Sigma=LL^T,
\end{equation}
where $L \in \mathbb{R}^{2 \times 2}$ is a lower triangular matrix with three independent covariance parameters, denoted as $L = (l_1, l_2, l_3)$.

Considering the absence of depth information in 2D images, GSImage replaces $\alpha$-blending rasterization with accumulated blending-based rasterization:
\begin{equation}
C_k=\sum_{i\in \mathcal{N}}C_i'\cdot\exp(-\sigma_i),\quad
\sigma_i=\frac{1}{2}d_i^T\Sigma^{-1}d_i.
\end{equation}

Thus, each 2D Gaussian is represented by 8 parameters: position vector $\mu \in \mathbb{R}^2$, covariance parameters $L \in \mathbb{R}^3$, and color coefficients $c' \in \mathbb{R}^3$.

To achieve compression, GSImage applies residual vector quantization to the color coefficients $c'$, applies 6-bit learned scale quantization to the covariance parameters $L$ and stores the position vector $\mu$ with 16-bit precision.

\subsection{Structure-guided Initialization}\label{SGI}
At initialization, GSImage randomly places 2D Gaussians within the normalized image domain $[-1,1]\times[-1,1]$ without leveraging structural priors. While adequate with a large number of Gaussians, this strategy is suboptimal under a limited budget, as many 2D Gaussians are wasted on low-detail regions while complex areas receive insufficient coverage.

\begin{figure}[!t]
\centering
\includegraphics[width=3.2in]{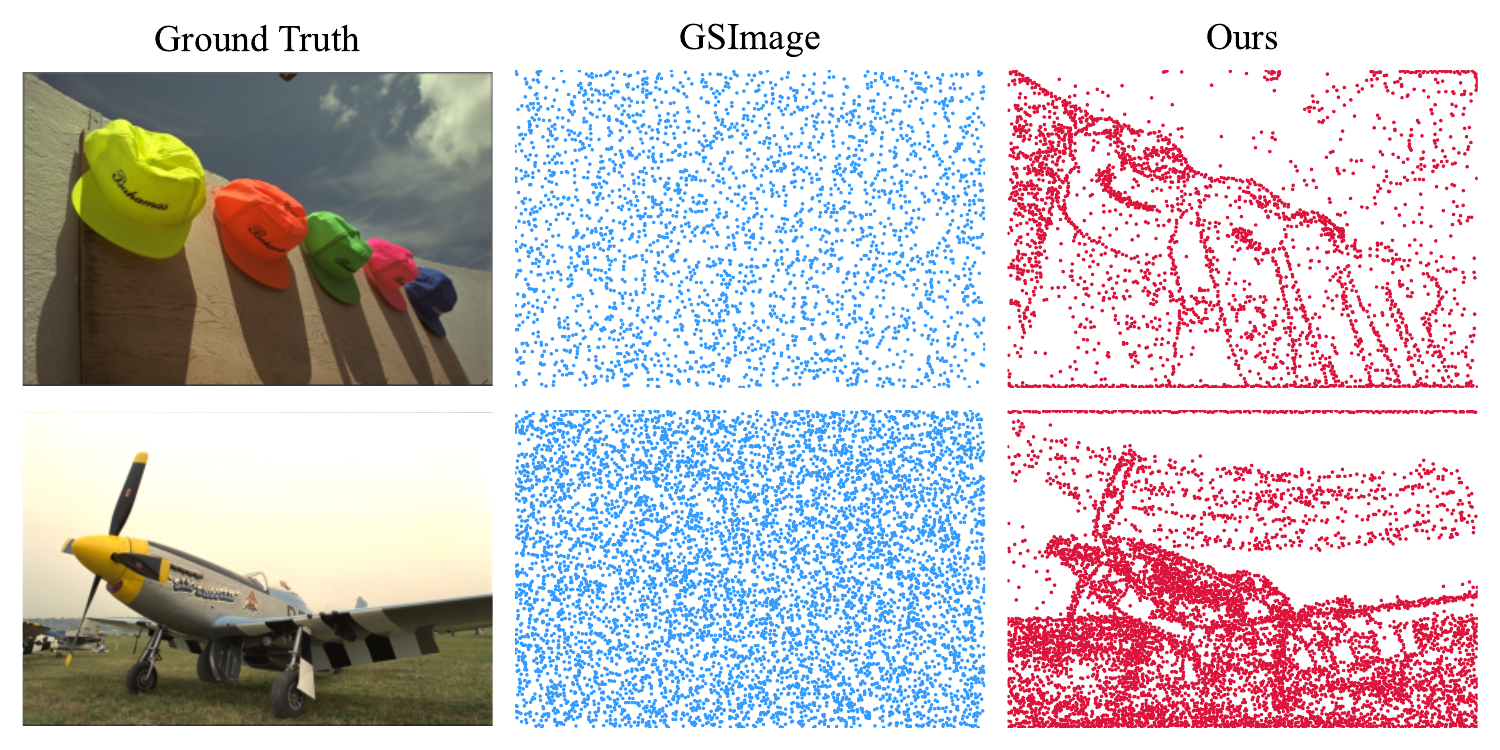}
\caption{Visualization of initial 2D Gaussians‘ distribution. Our method adaptively allocates more Gaussians to complex regions. The first row shows kodim03 (3k Gaussians) and the second row shows kodim20 (7k Gaussians)}
\label{Subject_Init_Compaison}
\end{figure}

\begin{figure}[!t]
\centering
\includegraphics[width=3.0in]{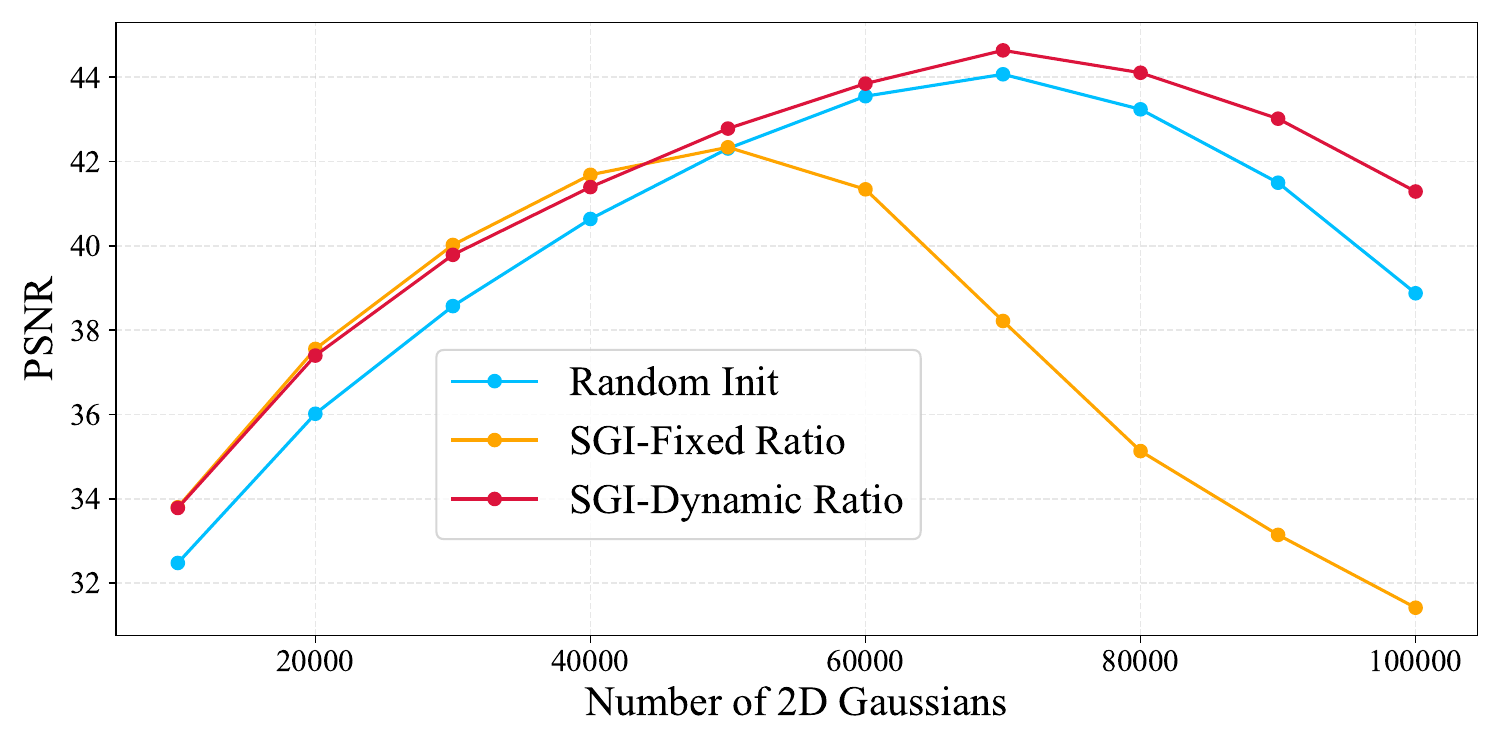}
\caption{Performance comparison of various initialization methods on the Kodak dataset, with SGI denoting our proposed Structure-Guided Initialization.}
\label{Comparison_Init}
\end{figure}

\begin{figure*}[!t]
\centering
\includegraphics[width=7.0in]{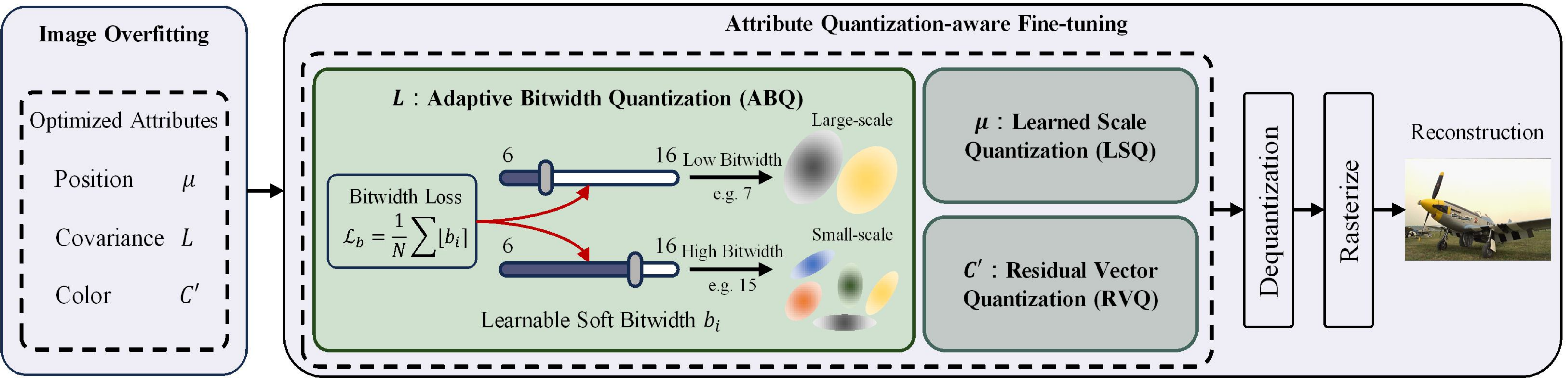}
\caption{Compression pipeline of the proposed method. After image overfitting, quantization-aware fine-tuning is performed to remove parameter redundancy and enable efficient compression. We introduce adaptive bitwidth quantization for covariance parameters, allocating higher precision to small-scale Gaussians in complex regions while using lower precision elsewhere. In addition, learned scale quantization is applied to position vector, and residual vector quantization is employed for color coefficients.}
\label{Compression_Pipeline}
\end{figure*}

\begin{figure}[!t]
\centering
\includegraphics[width=3.35in]{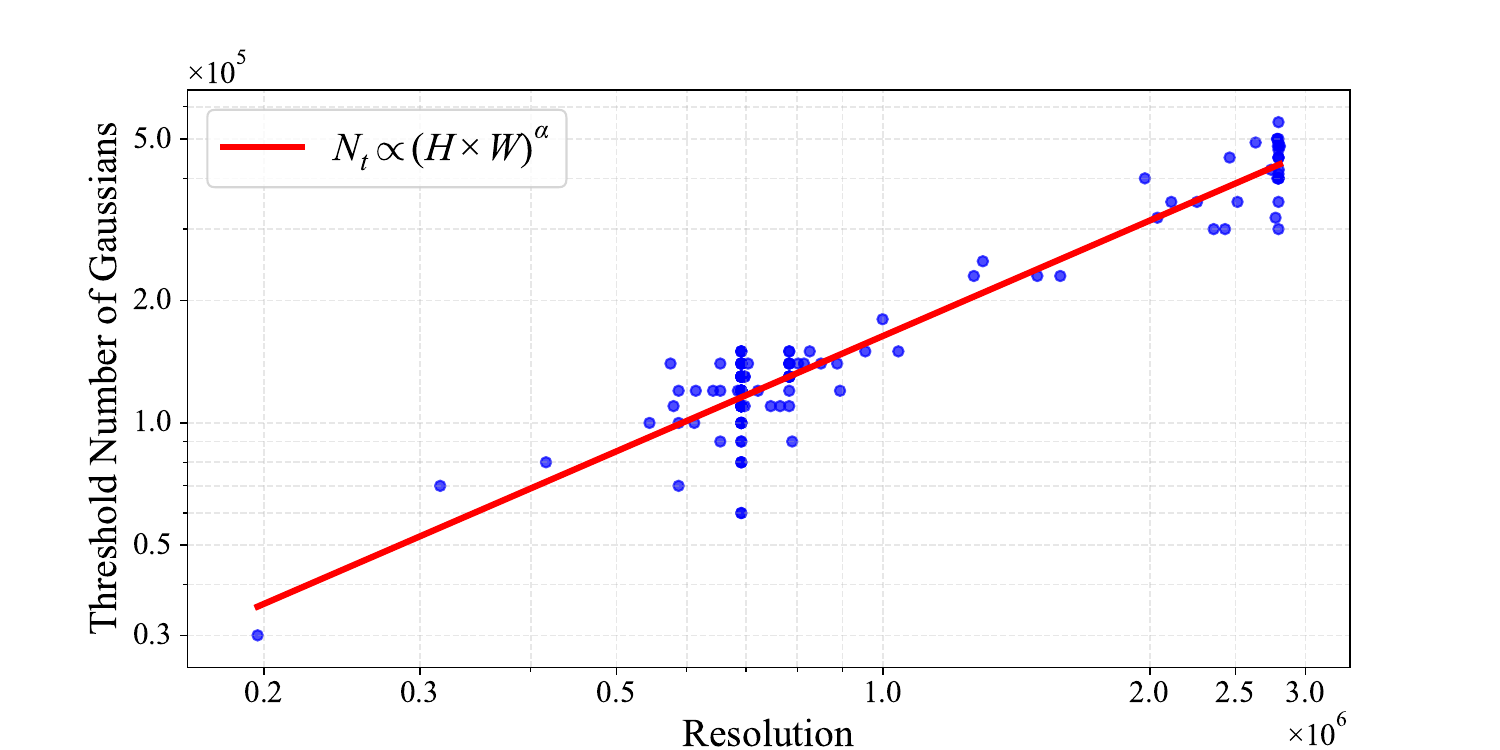}
\caption{Relationship between image resolution and the threshold number $N_t$ of 2D Gaussians, illustrating a sublinear scaling.}
\label{Resolution_number}
\end{figure}

To address this, we propose a structure-guided initialization, which adaptively allocates more Gaussians to regions with rich textures and complex structures (as shown in Fig.~\ref{Subject_Init_Compaison}), thereby improving rendering quality. We first convert the input RGB image to grayscale and compute its horizontal and vertical Sobel gradients, from which we derive the gradient magnitude map $G$. Next, the original RGB image is segmented using the SLIC algorithm~\cite{SLIC} to obtain a segmentation map, and the gradient map $G$ is partitioned accordingly. The variance of each region in $G$ is calculated to measure local structural complexity. Based on this metric, regions are sorted in descending order and divided into three categories. 2D Gaussians are allocated among these categories according to predefined ratios ($\phi_1:\phi_2:\phi_3,\phi_1\geq\phi_2\geq\phi_3$) and uniformly assigned within each category, with remaining points sequentially allocated to the most complex regions. This strategy ensures that more Gaussians are allocated to complex areas.

As illustrated in Fig.~\ref{Comparison_Init}, when the number of Gaussians exceeds 50,000, using a fixed ratio leads to a performance drop. This occurs because, at higher point counts, the optimal allocation tends toward a uniform distribution across all regions. In contrast, a fixed ratio overemphasizes complex regions, resulting in insufficient coverage of smoother areas, which limits representational capacity.

To overcome this, we introduce a dynamic ratio allocation, where the ratios $(\phi_1,\phi_2,\phi_3)$ vary with the total number $n$ of 2D Gaussians. When $n$ exceeds the threshold number $N_t$, the representational capacity of 2D Gaussians begins to degrade, primarily because the lack of depth information causes interference among nearby points. As shown in Fig.~\ref{Resolution_number}, when the resolution rises from 0.6M to 2.8M ($\approx4.6\times$), the required $N_t$ increases by only about $4\times$, indicating a sublinear relationship between the threshold $N_t$ and the resolution $H\times W$:

\begin{equation}
    N_t=k\cdot (H\times W)^\alpha,\,\alpha<1.
\end{equation}

Thus, we use $N_t$ to define the dynamic ratio for $n \leq N_t$ as:
\begin{equation}
\begin{aligned}
    \phi_i(n)=(1-s(n))\cdot\phi_i+s(n)\cdot\frac{1}{3},\,i=1,2,3,
\end{aligned}\label{Dynamic_Ratio}
\end{equation}
where $s(n)=\text{clamp}\left(\left(\frac{n-n_0}{N_t-n_0}\right)^\gamma;0,1\right)$ smoothly shifts the allocation from $(\phi_1:\phi_2:\phi_3)$ toward a uniform $(1:1:1)$ as $n$ approaches the threshold $N_t$. As shown in Fig.~\ref{Comparison_Init}, the dynamic ratio allocation consistently improves reconstruction quality across a broad range of point budgets.

\subsection{Adaptive Bitwidth Quantization}\label{ABQ}
Following overfitting training, a compression pipeline is needed to reduce model size and enable efficient compression (as shown in Fig.~\ref{Compression_Pipeline}). In GSImage, the covariance parameters $L$ are uniformly quantized using 6-bit learned scale quantization. Such uniform bitwidth allocation is suboptimal, as small-scale Gaussians in complex regions suffer noticeable errors, degrading RD performance. To address this, we propose an adaptive bitwidth quantization strategy that assigns higher precision to small-scale Gaussians and lower precision elsewhere, in accordance with the structure-guided allocation principle. This design effectively enables RD optimization, improving compression efficiency while maintaining reconstruction fidelity.

Given an input value $v$ to be quantized, the learned scale quantization introduces a learnable step size $s$ and a learnable zero-point offset $Z$. The quantization and dequantization processes are defined as follows:
\begin{equation}
    \hat{v}=s\cdot\tilde{v}+Z=s\cdot\text{clamp}\left(\left\lfloor \frac{v-Z}{s}\right\rceil;q_{\min},q_{\max}\right)+Z,
\end{equation}
where $\lfloor \cdot \rceil$ denotes rounding to the nearest integer and $\operatorname{clamp}(\cdot; q_{\min}, q_{\max})$ restricts the quantized value to the range $[q_{\min},q_{\max}]$. For a bitwidth precision $B$, the representable integer range satisfies $Q=q_{\max}-q_{\min}=2^{B}-1$. The step size $s$ and zero-point offset $Z$ are given by:
\begin{equation}
    Z = v_{\min}, \,s=\frac{v_{\max}-v_{\min}}{2^B-1}.
\end{equation}

To make the bitwidth learnable, we introduce a soft bitwidth $b$ in adaptive bitwidth quantization and derive a hard bitwidth $B=\lfloor b \rceil$. During backpropagation, the straight-through estimator (STE) is used to approximate the gradient of the rounding operation. Consequently, $v_{\max}$, $v_{\min}$, and $b$ are treated as learnable parameters during training. Under the STE approximation, their gradients for backpropagation are derived as follows:

\begin{equation}
\begin{aligned}
\frac{\partial \hat{v}}{\partial v_{\max}} &= 
\begin{cases}
\dfrac{\tilde{v}}{Q} - \dfrac{v-v_{\min}}{sQ}, & v_{\min} \leq v \leq v_{\max}, \\
\dfrac{\tilde{v}}{Q}, & v < v_{\min} \ \text{or}\ v > v_{\max},
\end{cases} \\[6pt]
\frac{\partial \hat{v}}{\partial v_{\min}} &= 
\begin{cases}
-\dfrac{\tilde{v}}{Q}+\dfrac{v-v_{\min}}{sQ}, & v_{\min} \leq v \leq v_{\max}, \\
1-\dfrac{\tilde{v}}{Q}, & v < v_{\min} \ \text{or}\ v > v_{\max},
\end{cases} \\[6pt]
\frac{\partial \hat{v}}{\partial b} &= 
\begin{cases}
\dfrac{2^B\ln 2}{Q}\left(v-\tilde{v}\right), & v_{\min} \leq v \leq v_{\max}, \\
\dfrac{2^B\ln 2}{Q}\left(-s\tilde{v}\right), & v < v_{\min} \ \text{or}\ v > v_{\max}.
\end{cases}
\end{aligned}
\end{equation}

In the learned scale quantization, $s$ and $Z$ are learned by minimizing the MSE loss. However, optimizing solely via the MSE loss does not produce bitwidths that maximize compression efficiency, as an MSE-only objective tends to favor larger bitwidths to reduce quantization error. To counteract this tendency, we introduce a bitwidth loss term that explicitly penalizes large bitwidths and enforces RD optimization. Formally, the bitwidth loss can be written as:
\begin{equation}
\mathcal{L}_{b} = \frac{1}{\mathcal{N}}\sum_{i\in\mathcal{N}} B_i,
\end{equation}
where $B_i = \lfloor b_i \rceil$ is the hard bitwidth of the $i$-th 2D Gaussian.

During encoding, 2D Gaussians are grouped by their hard bitwidths $B$, and only the bitwidth of each group is stored as metadata, rather than storing each Gaussian’s bitwidth individually. Therefore, the adaptive bitwidth quantization scheme incurs extra metadata overhead ($\approx$ 0.0004 bpp on Kodak) and results in only a negligible reduction in decoding speed.

For the position vector, since each parameter contributes equally to reconstruction quality, we use a uniform bitwidth with learned scale quantization for all Gaussians, ensuring accurate reconstruction while reducing redundancy.

\begin{table*}[!t]
\caption{Quantitative comparison on the Kodak dataset with various baselines in PSNR, MS-SSIM, LPIPS, FID, training-time, rendering speed, GPU memory usage and parameter size.\label{Quantitative_Evaluation_Table1}}
\centering
\scriptsize
\begin{adjustbox}{width=\textwidth}
\begin{tabular}{lcccccccc}
\toprule
Method & PSNR $\uparrow$ & MS-SSIM $\uparrow$ & LPIPS $\uparrow$ & FID $\uparrow$ &Training Time (s) $\downarrow$ & FPS $\uparrow$ & GPU Mem. (MiB) $\downarrow$ & Params (K) $\downarrow$ \\
\midrule
SIREN~\cite{SIREN} & 40.83 & 0.9960 & 0.0474 & 4.6873 & 6582.36 & 29.15 & 1809 & {\cellcolor{yellow!40} 272.70}\\
WIRE~\cite{WIRE} & 41.47 & 0.9939 & 0.0317 & 3.2130 & 14338.78 & 11.14 & 2619 & {\cellcolor{orange!40} 136.74}\\
I-NGP~\cite{Instant-NGP} & 43.88 & 0.9976 & 0.0171 & 1.2518 & 490.61 & 1296.82 & 1525 & 300.09\\
NeuRBF~\cite{Neurbf} & 43.78 & 0.9964 & 0.0191 & 1.8623 & 991.83 & 663.01 & 2091 & 337.29\\
3D-GS~\cite{3DGS} & 43.69 & 0.9991 & 0.0219 & 2.4893 & 339.78 & 859.44 & 557 & 3540.00\\
GSImage~\cite{Gaussianimage} & {\cellcolor{yellow!40} 44.08} & {\cellcolor{yellow!40} 0.9985} & {\cellcolor{yellow!40} 0.0083} & {\cellcolor{yellow!40} 0.9120} & {\cellcolor{orange!40} 106.59} & {\cellcolor{yellow!40} 2092.17} & {\cellcolor{orange!40} 419} & 560.00\\
\midrule
Ours & {\cellcolor{orange!40} 45.40} & {\cellcolor{orange!40} 0.9987} & {\cellcolor{orange!40} 0.0047} & {\cellcolor{orange!40} 0.6432} & {\cellcolor{yellow!40} 134.42} & {\cellcolor{orange!40} 2109.05} & {\cellcolor{yellow!40} 484} & 560.00\\
\bottomrule
\end{tabular}
\end{adjustbox}
\end{table*}

\begin{table*}[ht]
\caption{Quantitative evaluation on the DIV2K $\times 2$ dataset. \label{Quantitative_Evaluation_Table2}}
\centering
\scriptsize
\begin{adjustbox}{width=\textwidth}
\begin{tabular}{lcccccccc}
\toprule
Method & PSNR $\uparrow$ & MS-SSIM $\downarrow$ & LPIPS $\downarrow$& FID $\uparrow$ & Training Time (s) $\downarrow$ & FPS $\uparrow$ & GPU Mem. (MiB) $\downarrow$ & Params (K) $\downarrow$ \\
\midrule
SIREN~\cite{SIREN} & 39.08 & 0.9958 & {\cellcolor{orange!40} 0.0111} & 2.5894 & 15125.11 & 11.07 & 2053 & 483.60\\
WIRE~\cite{WIRE} & 35.64 & 0.9511 & 0.0281 & 9.7842 & 25684.23 & 14.25 & 2619 & {\cellcolor{orange!40} 136.74}\\
I-NGP~\cite{Instant-NGP} & 37.06 & 0.9950 & 0.0218 & 4.6752 & 676.29 & 1331.54 & 1906 & 525.40\\
NeuRBF~\cite{Neurbf} & 38.60 & 0.9913 & 0.0126 & 2.8764 & 1715.44 & 706.40 & 2893 & {\cellcolor{yellow!40} 383.65}\\
3D-GS~\cite{3DGS} & 39.36 & 0.9979 & {\cellcolor{yellow!40} 0.0119} & 2.5972 & 481.27 & 640.33 & 709 & 4130.00\\
GSImage~\cite{Gaussianimage} & {\cellcolor{yellow!40} 39.53} & {\cellcolor{orange!40} 0.9975} & 0.0131 & {\cellcolor{yellow!40} 2.5607} & {\cellcolor{orange!40} 120.76} & {\cellcolor{yellow!40} 1737.60} & {\cellcolor{orange!40} 439} & 560.00\\
\midrule
Ours & {\cellcolor{orange!40}41.29} & {\cellcolor{yellow!40}0.9974} & 0.0159 & {\cellcolor{orange!40} 2.2416} & {\cellcolor{yellow!40} 172.11} & {\cellcolor{orange!40} 1837.77} & {\cellcolor{yellow!40} 646} & 560.00\\
\bottomrule
\end{tabular}
\end{adjustbox}
\end{table*}

\subsection{Geometry-consistent Regularization}\label{GCR}
To further facilitate structure-guided optimization, we propose a geometry-consistent regularization that aligns 2D Gaussians with local gradient orientations to better preserve fine structural details. Unlike conventional MSE loss, which treats all pixels equally, our regularization assigns higher importance to regions with greater structural complexity, thereby promoting better preservation of fine structures and textures. To this end, we leverage a gradient-based loss to emphasize regions with higher visual complexity. Specifically, we employ Sobel operators to extract the horizontal and vertical gradient maps $({G'}_x,{G'}_y,G_x,G_y)$ for both the reconstructed image $I'\in\mathbb{R}^{H\times W\times C}$ and its ground-truth image $I$.

We then compute the element-wise squared differences between the gradient maps of the reconstructed image and the ground-truth image $(I',I)$ to obtain the horizontal and vertical gradient discrepancy maps $(D_x, D_y)$:
\begin{equation}\label{gradient_map}
\begin{aligned}
D_x=(G'_x-G_x)^2,\, D_y=(G'_y-G_y)^2.
\end{aligned}
\end{equation}

A straightforward approach is to average the two discrepancy maps $(D_x, D_y)$ as the final loss. However, this leads to a critical issue: in images dominated by flat regions with low gradient magnitudes, the loss overemphasizes reconstruction accuracy in these uninformative areas. This behavior conflicts with the intent of geometry-consistent supervision, which aims to guide the model toward structurally complex regions and preserve fine textures. To address this, we introduce geometry-consistent weighting terms $(W_x, W_y)$, defined as the magnitudes of the ground-truth gradients, to amplify the contribution of structure-rich areas:
\begin{equation}\label{Gradient_Weight}
W_x = |G_x|, \quad W_y = |G_y|.
\end{equation}

These weights assign greater importance to visually complex regions, guiding the model to focus on perceptually significant details. The final geometry-consistent regularization term is formulated as:
\begin{equation}
\mathcal{L}_g = \frac{1}{HW}\sum_{i=1}^{H}\sum_{j=1}^{W}\left(W_xD_x + W_yD_y\right).
\end{equation}

\subsection{Optimization}\label{Optimization}
For the image representation task, our objective is to minimize the discrepancy between the reconstructed image and the ground-truth image, thereby ensuring high-fidelity reconstruction. A key challenge lies in striking a balance between preserving the global structure and recovering fine-grained local details. To address this, we adopt a weighted combination of the MSE loss $\mathcal{L}_{MSE}$ and the geometry-consistent regularization $\mathcal{L}_g$ as the training objective function, formulated as follows:
\begin{equation}
\mathcal{L}_{train}=\mathcal{L}_{MSE}+\lambda_g\mathcal{L}_g\, ,
\end{equation}
where the hyperparameter $\lambda_g$ controls the trade-off between global fidelity and local detail. Increasing $\lambda_g$ directs more capacity to complex, high-frequency regions, preserving texture while potentially sacrificing global reconstruction accuracy.

For image compression, the fine-tuning loss $\mathcal{L}_{tune}$ consists of the reconstruction loss $\mathcal{L}_{train}$, bitwidth loss $\mathcal{L}_{b}$, and residual compensation loss $\mathcal{L}_r$:
\begin{equation}
    \mathcal{L}_{tune} = \mathcal{L}_{train} + \lambda_b \mathcal{L}_b + \lambda_r \mathcal{L}_r\, ,
\end{equation}
where $\mathcal{L}_{b}$ is the bitwidth penalty that encourages compact representations by penalizing large bitwidth allocations, and $\mathcal{L}_r$ measures the discrepancy between the quantized color attributes produced by residual vector quantization and their corresponding ground-truth values. Optimizing the tuning loss $\mathcal{L}_{tune}$ jointly adjusts the bitwidths, quantization parameters, and model representations in a coordinated manner, thereby achieving RD-aware optimization.

\begin{figure*}[!ht]
\centering
\includegraphics[width=7.0in]{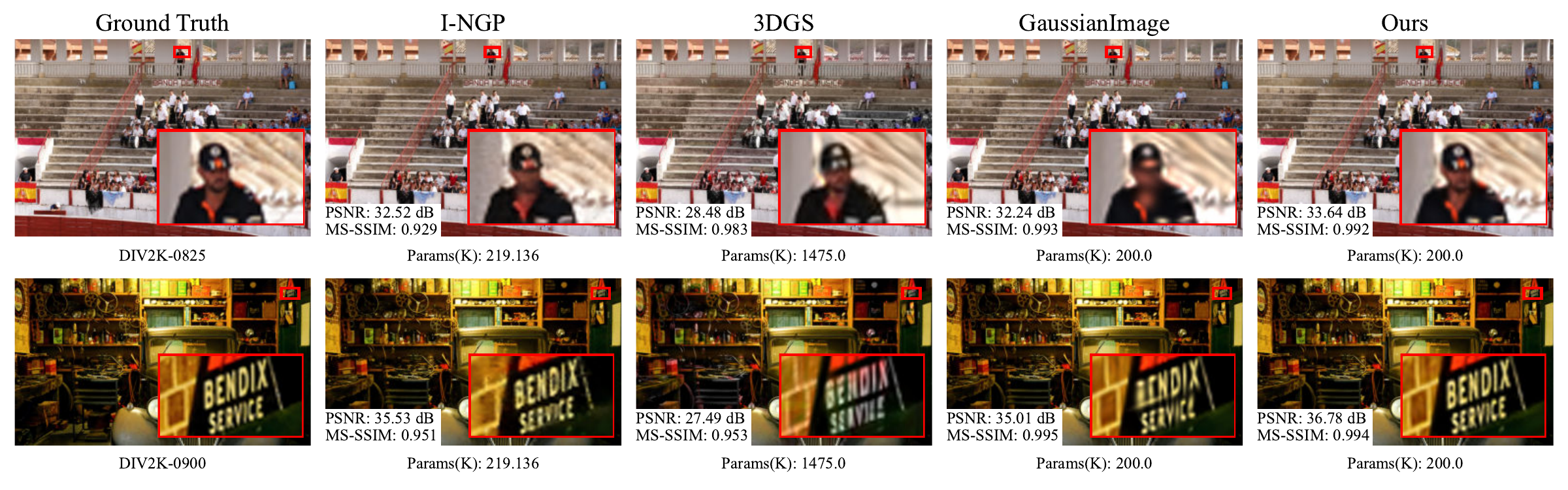}
\caption{Qualitative comparison of our approach with various baselines on the DIV2K $\times 2$ dataset. The leftmost column shows the ground-truth images, while the remaining columns display the reconstructed outputs of each method.}
\label{Qualitative_Comparison_DIV2K}
\end{figure*}

\begin{figure*}[!t]
\centering
\includegraphics[width=6.0in]{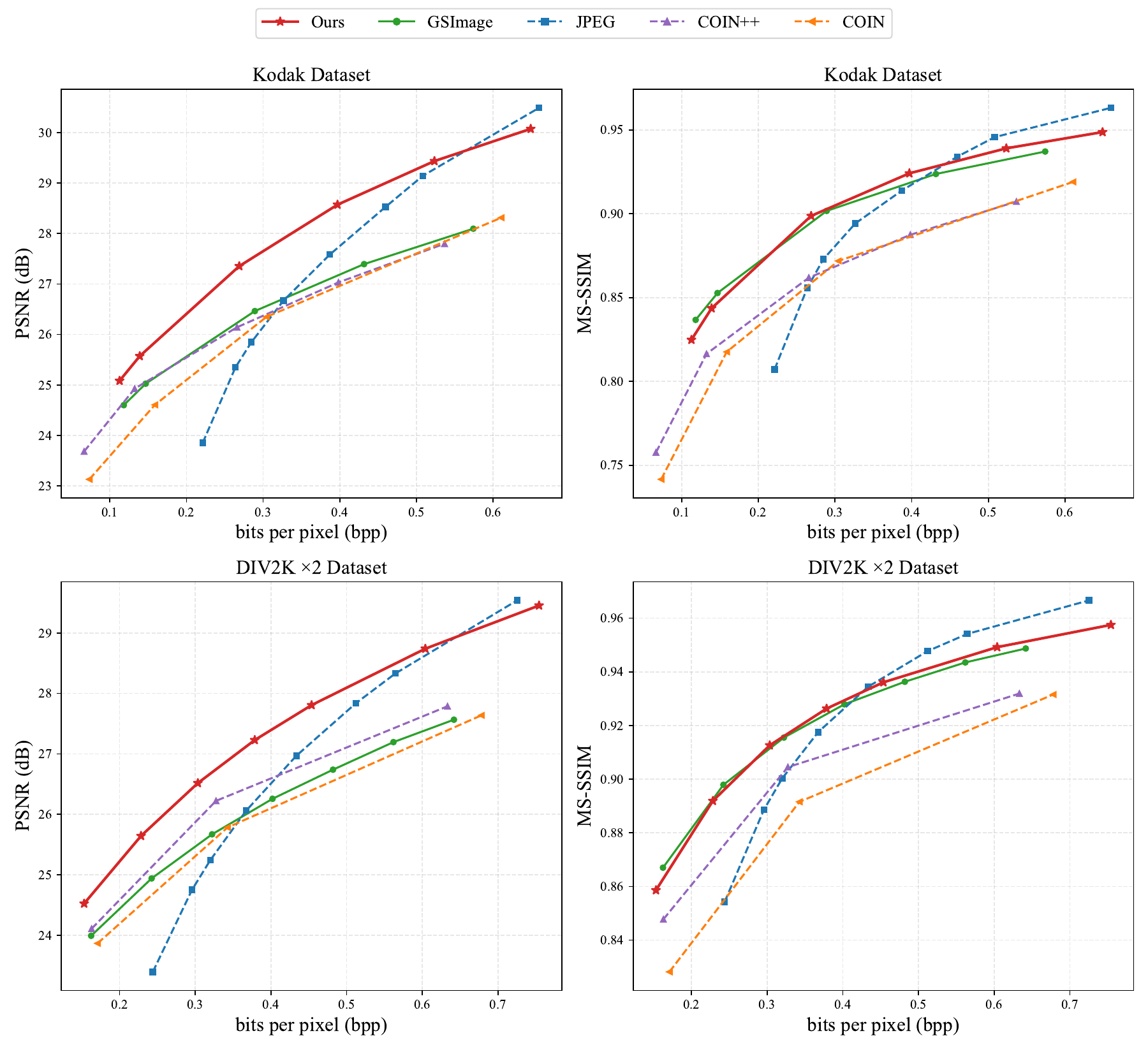}
\caption{RD curves of our approach and different baselines on the Kodak and DIV2K $\times 2$ datasets in PSNR and MS-SSIM.}
\label{RD_Curve_Comparison}
\end{figure*}

\section{Experimental Results}
\subsection{Experimental Settings}
\textbf{Dataset.} We evaluate our method on both the image representation and compression tasks using three benchmark datasets: Kodak, DIV2K~\cite{DIV2K}, and CLIC2020 validation sets. For DIV2K, we adopt two settings: DIV2K (original high-resolution images) and DIV2K $\times 2$ (bicubic downsampled images). The diverse content and resolutions of DIV2K and CLIC2020 provide a comprehensive basis for assessing the generalization performance of different methods.

\textbf{Evaluation Metrics.} To evaluate representation performance, we use PSNR, MS-SSIM~\cite{MS-SSIM}, and LPIPS~\cite{LPIPS} as quality metrics. For image compression, the bitrate is measured in bpp. Additionally, BD-rate and BD-PSNR~\cite{BD-Rate} are employed to compare overall RD performance across methods.

\textbf{Implementation Details.} The predefined ratios are set to $\phi_1:\phi_2:\phi_3=6:2:1$, with $\gamma=10.0$ and $n_0=10,000$ in Eq.~\ref{Dynamic_Ratio}. The weight of the geometry-consistent regularization term $\lambda_g$ is set to 0.06. During quantization-aware fine-tuning, the quantization settings for color attributes follow those used in GSImage. For the adaptive bitwidth quantization (ABQ) scheme, the bitwidth loss weight is 0.0012, and the covariance parameters are constrained to the range [6, 16] bits. The position vectors are quantized using 12-bit learned scale quantization. All experiments are conducted in PyTorch on NVIDIA RTX 4090 GPUs.

\textbf{Benchmarks.} For image representation comparisons, our method is evaluated against several state-of-the-art implicit neural representation (INR) methods, including SIREN~\cite{SIREN}, WIRE~\cite{WIRE}, I-NGP~\cite{Instant-NGP}, and NeuRBF~\cite{Neurbf}, as well as the baseline GSImage~\cite{Gaussianimage}. For image compression, the comparison includes traditional codecs (JPEG~\cite{JPEG}, JPEG2000~\cite{JPEG2000}), INR-based codecs (COIN~\cite{COIN}, COIN++~\cite{COIN++}), and 2DGS-based approaches (2DGS~\cite{2DGS}, GSImage~\cite{Gaussianimage}). All INR methods are trained with the same number of optimization steps as our method to ensure a fair comparison.

\subsection{Image Representation}
\textbf{Quantitative Comparison.} Tables~\ref{Quantitative_Evaluation_Table1} and~\ref{Quantitative_Evaluation_Table2} summarize the quantitative evaluation results on the Kodak and DIV2K $\times 2$ datasets under identical optimization settings. Compared with both MLP-based and grid-based baselines, our method achieves the best overall trade-off among reconstruction quality, computational efficiency, and GPU memory usage, demonstrating its superior representational efficiency. Built upon GSImage, the proposed structure-guided initialization and geometry-consistent regularization jointly enhance the effective utilization of representational capacity, leading to the highest reconstruction accuracy. Specifically, our method attains PSNR values of 45.40 dB on Kodak and 41.29 dB on DIV2K $\times 2$, while maintaining competitive training speed (exceeding 1,000 FPS), inference efficiency, and memory consumption.

\textbf{Qualitative Comparison.} Fig.~\ref{Qualitative_Comparison_DIV2K} presents qualitative reconstruction results on the DIV2K $\times 2$ dataset. Compared with competing approaches under the same optimization steps, our method consistently preserves finer structural and textural details, especially in regions with complex structure or high-frequency content. For example, on DIV2K-0900, our method achieves the highest PSNR of 36.78 dB while maintaining a comparable parameter count, and accurately reconstructs fine text-like details (e.g., characters and edges) with noticeably fewer visual artifacts. These results highlight the strong representational capacity of our model in capturing fine-grained textures and structural details.

\begin{table*}[!ht]
\caption{Computational complexity of various codecs on the DIV2K $\times 2$ dataset at low and high bitrates.\label{Computation_Complexity_Table}}
\centering
\scriptsize
\begin{adjustbox}{width=\textwidth}
\begin{tabular}{lcccccccccccc}
\toprule
\multirow{2}{*}{Method} & \multicolumn{5}{c}{Low bitrate} & \multicolumn{5}{c}{High bitrate} \\
\cmidrule(lr){2-6} \cmidrule(lr){7-11}
 & bpp & PSNR & MS-SSIM & Encoding FPS & Decoding FPS & bpp & PSNR & MS-SSIM & Encoding FPS & Decoding FPS\\
\midrule
JPEG~\cite{JPEG} & 0.3197 & 25.29 & 0.9020 & 608.61 & 614.68 & 0.5638 & 28.43 & 0.9559 & 557.35 & 545.59\\
JPEG2000~\cite{JPEG2000} & 0.2394 & 27.28 & 0.9305 & 3.46 & 4.32 & 0.5993 & 30.93 & 0.9663 & 3.40 & 3.93\\
COIN~\cite{COIN} & 0.3419 & 25.80 & 0.8905 & $5.30e^{-4}$ & 166.31 & 0.6780 & 27.61 & 0.9306 & $3.51e^{-4}$ & 93.74\\
2DGS~\cite{2DGS} & 0.0420 & 20.74 & 0.6206 & $3.89e^{-3}$ & 54.752 & 0.1830 & 23.99 & 0.8305 & $4.09e^{-3}$ & 52.346 \\
GSImage~\cite{Gaussianimage} & 0.3221 & 25.66 & 0.9154 & $4.11e^{-3}$ & 1970.76 & 0.6417 & 27.57 & 0.9483 & $4.73e^{-3}$ & 1980.54\\
\midrule
Ours & 0.3033 & 26.52 & 0.9126 & $6.26e^{-3}$ & 1540.87 & 0.6039 & 28.74 & 0.9492 & $6.92e^{-3}$ & 1519.39\\
\bottomrule
\end{tabular}
\end{adjustbox}
\end{table*}

\begin{figure}[!t]
\centering
\includegraphics[width=3.5in]{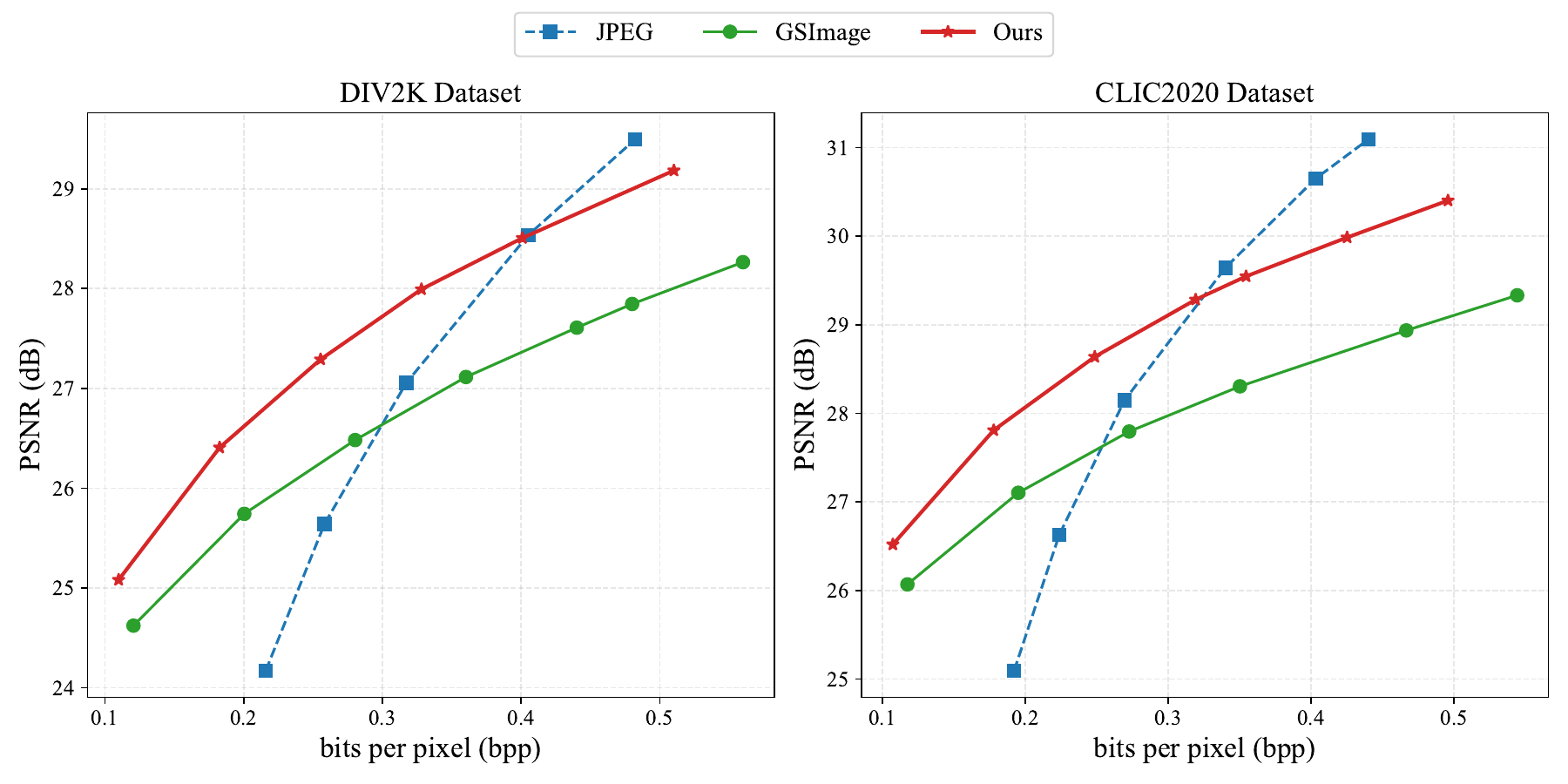}
\caption{RD curves on the high resolution datasets DIV2K and CLIC2020.}
\label{RD_Curve_HR_Comparison}
\end{figure}


\subsection{Image Compression}
\textbf{Compression Performance.} Fig.~\ref{RD_Curve_Comparison} illustrates the RD curves of various codecs evaluated on the Kodak and DIV2K $\times 2$ datasets. The results clearly demonstrate that our method surpasses the baseline GSImage in both PSNR and MS-SSIM. Specifically, our method achieves BD-rate reductions of 43.44\% on the Kodak dataset and 29.91\% on the DIV2K $\times 2$ dataset compared with GSImage, highlighting the effectiveness of our proposed method in improving compression efficiency. Notably, as shown in Fig.~\ref{RD_Curve_Comparison} and Fig.~\ref{RD_Curve_HR_Comparison}, our method achieves better RD performance than the traditional JPEG codec at low bitrates (e.g., below 0.4 bpp). Furthermore, as illustrated in Fig.~\ref{Compression_Comparison}, our approach not only delivers superior RD performance but also maintains faster decoding speeds than JPEG, demonstrating its practical suitability for real-time applications.

\textbf{Computational Complexity.} Fig.~\ref{Compression_Comparison} and Table~\ref{Computation_Complexity_Table} summarize the computational complexity of various codecs on the Kodak and DIV2K $\times 2$ datasets. Our method achieves an impressive decoding speed of 1,700 FPS on Kodak and 1,500 FPS on DIV2K while simultaneously surpassing GSImage in compression efficiency. Notably, the compression performance of our method surpasses that of JPEG at low bitrates while delivering faster decoding speed. The efficiency stems from structure-guided initialization and geometry-consistent regularization, which are applied only during training, and adaptive bitwidth quantization, which reduces parameter redundancy and has minimal impact on inference speed.

\begin{figure}[!t]
\centering
\includegraphics[width=3.0in]{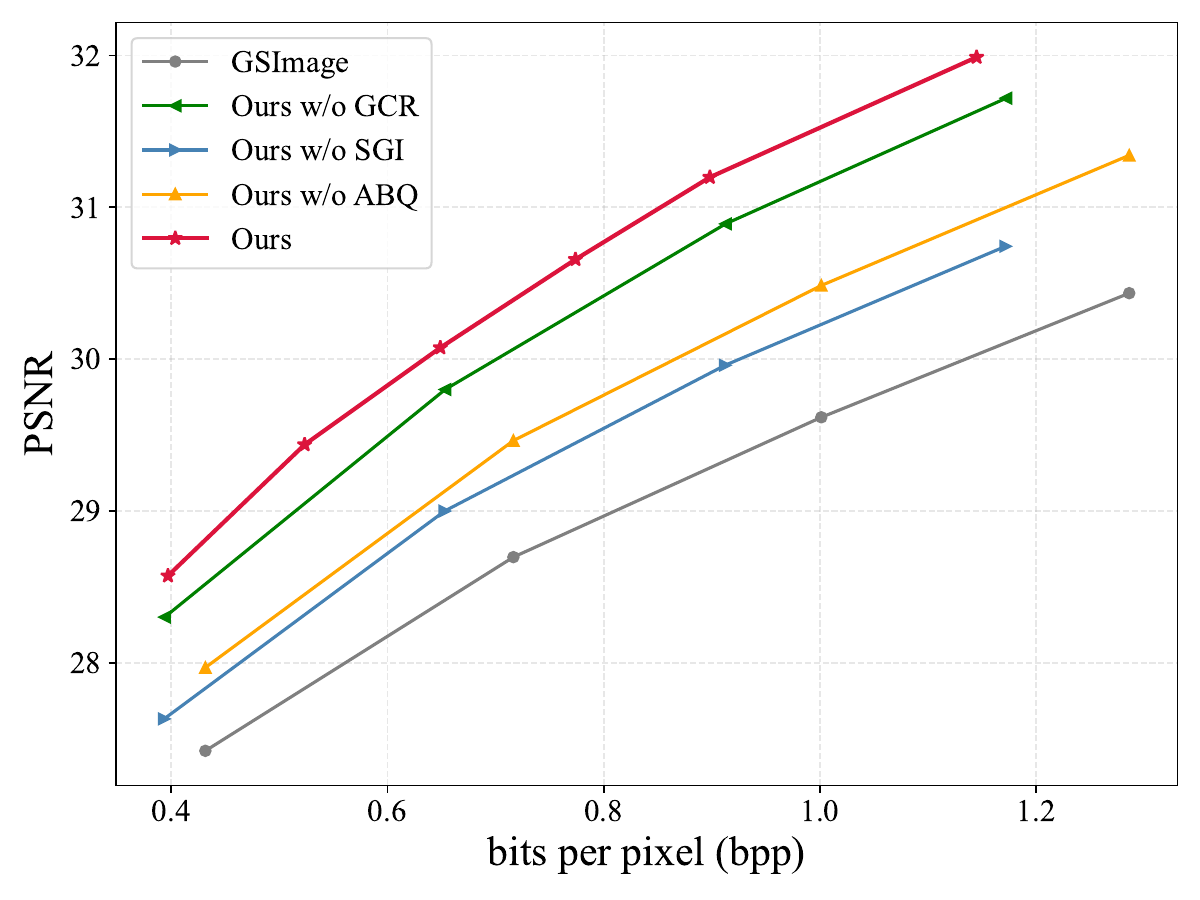}
\caption{RD curves of Geometry-Consistent Regularization (GCR), Structure-Guided Initialization (SGI), and Adaptive Bitwidth Quantization (ABQ) on the Kodak dataset in PSNR.}
\label{RD_Curve_Ablation}
\end{figure}

\begin{figure*}[!t]
\centering
\includegraphics[width=7.0in]{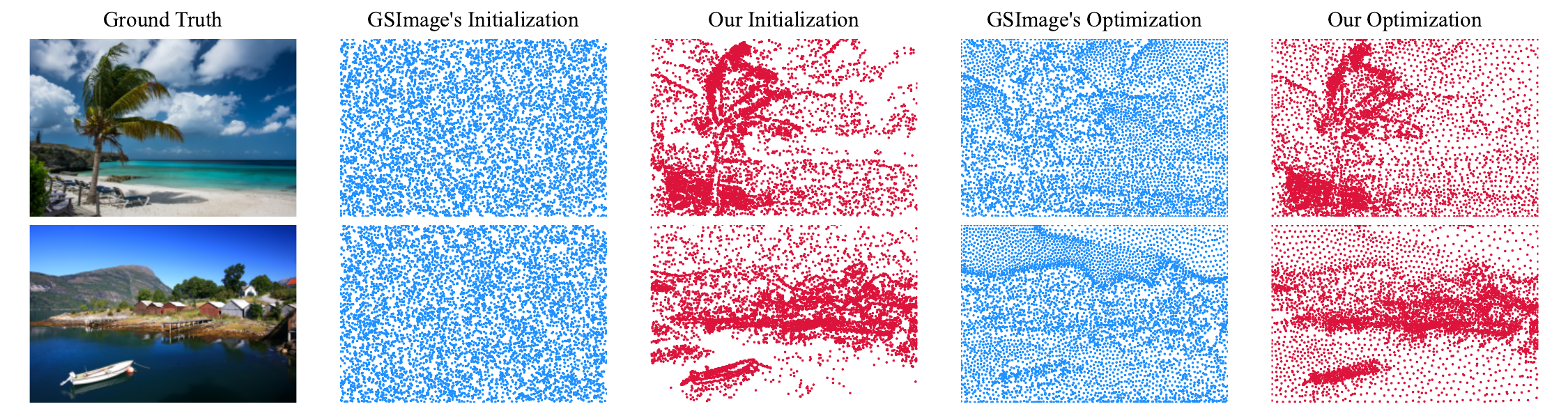}
\caption{Visual comparison of the initial and optimized distributions of 2D Gaussians between GSImage and our method, both using 6,000 Gaussians on DIV2K $\times 2$ dataset, highlighting the structural improvement achieved by our approach.}
\label{Init_Optimized_Point_cloud}
\end{figure*}

\begin{table}[!t]
\caption{Ablation study of Geometry-Consistent Regularization (GCR) and Structure-Guided Initialization (SGI) in image representation on the Kodak dataset with 30,000 Gaussians.
\label{Ablation_Image_Representation}}
\centering
\begin{tabular}{lcccc}
\toprule
Method & PSNR & MS-SSIM & Training Time (s) & FPS \\
\midrule
GSImage & 38.57 & 0.9961 & 91.06 & 2283 \\
Ours (w/o GCR) & 40.02 & 0.9957 & 98.71 & 2193 \\
Ours (w/o SGI) & 38.75 & 0.9963 & 130.37 & 2071\\
Ours & 40.51 & 0.9958 & 131.94 & 2066\\
\bottomrule
\end{tabular}
\end{table}

\subsection{Ablation Results}
\textbf{Image Representation.} To evaluate the effectiveness of our method for image representation, we conducted comprehensive ablation studies summarized in Table~\ref{Ablation_Image_Representation}. Replacing random initialization with our structure-guided initialization (SGI) improves PSNR by 1.45 dB, demonstrating that SGI provides a more informative and semantically meaningful distribution for 2D Gaussians. Incorporating geometry-consistent regularization further increases PSNR by 0.50 dB by better preserving structural and edge details. As shown in Fig.~\ref{Init_Optimized_Point_cloud}, SGI initially concentrates 2D Gaussians in complex, detail-rich regions, and after optimization with MSE and regularization, the 2D Gaussians redistribute toward smoother regions while maintaining dense coverage in complex areas, ensuring both global fidelity and fine-grained reconstruction.

\textbf{Image Compression.} As shown in Fig.~\ref{RD_Curve_Ablation}, the structure-guided initialization improves encoder efficiency by enhancing reconstruction quality, achieving an 18\% BD-rate reduction over GSImage. When further combined with adaptive bitwidth quantization and learned scale quantization, which reduce redundancy in covariance parameters and position vectors, respectively, our method decreases the BD-rate by 43\% with a 1.6 dB BD-PSNR gain. Although geometry-consistent regularization has a limited impact on compression performance, as fine-grained details are often suppressed during quantization, its inclusion still improves the overall optimization.

\textbf{Geometry-consistent regularization.} We evaluate the weight $\lambda_g$ for the geometry-consistent regularization by sweeping values in the range from $0.02$ to $0.08$. On the Kodak dataset with 70,000 2D Gaussians, $\lambda_g=0.06$ produces the largest PSNR improvement, yielding a gain of approximately $0.42$ dB. This indicates that $\lambda_g=0.06$ effectively balances the emphasis on fine structural details and overall image fidelity. Therefore, $\lambda_g=0.06$ is adopted for our method.

\begin{table}[!t]
\caption{Ablation study on the weight $\lambda_g$ of geometry-consistent regularization (GCR) with 70,000 2D Gaussians on the Kodak dataset.
\label{GCR_Ablation}}
\centering
\begin{tabular}{lccc}
\toprule
Method & PSNR & MS-SSIM & LPIPS\\
\midrule
Ours w/o GCR & 44.98 & 0.99862 & 0.0061 \\
Ours w/ GCR ($\lambda_g=0.04$) & 45.36 & 0.99872 & 0.0050\\
Ours w/ GCR ($\lambda_g=0.06$) & 45.40 & 0.99874 & 0.0047\\
Ours w/ GCR ($\lambda_g=0.08$) & 45.38 & 0.99873 & 0.0047\\
\bottomrule
\end{tabular}%
\end{table}

\begin{table}[!t]
\small
\caption{Comparison of Learned Scale Quantization (LSQ) and Adaptive Bitwidth Quantization (ABQ) under our framework on the Kodak dataset, with BD-PSNR and BD-rate computed relative to the baseline GSImage.
\label{Ablation_ABQ}}
\centering
\resizebox{\columnwidth}{!}{
\begin{tabular}{lcc}
\toprule
Quantization Method of covariance & BD-PSNR (dB) & BD-rate (\%) \\
\midrule
LSQ (6-bit) & 0.7792 & -26.2867\\
LSQ (7-bit) & 1.1630 & -35.2255 \\
LSQ (8-bit) & 1.1766 & -35.3416 \\
\midrule
ABQ ($\lambda_b=0.0012$) & 1.6782 & -43.4365\\
ABQ ($\lambda_b=0.0014$) & 0.9596 & -30.2927\\
\bottomrule
\end{tabular}
}
\end{table}

\begin{figure}[!t]
\centering
\includegraphics[width=3.0in]{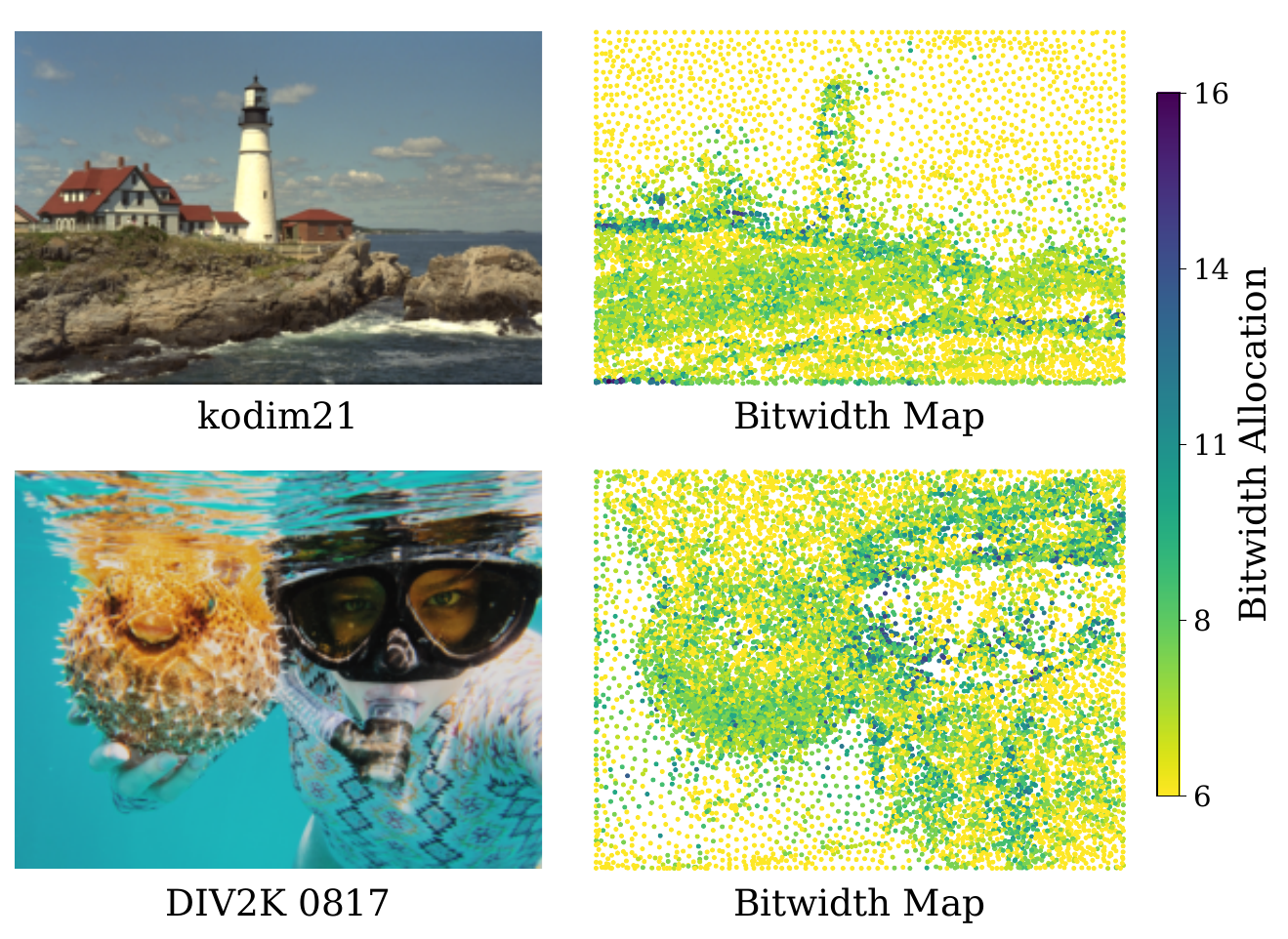}
\caption{Visualization of adaptive bitwidth allocation. Left: original image; right: bitwidth map generated by our adaptive bitwidth quantization.}
\label{Bitwidth_Map}
\end{figure}

\textbf{Adaptive bitwidth quantization.} To evaluate adaptive bitwidth quantization (ABQ), we quantize covariance parameters using different bitwidth loss weights $\lambda_b$ under our framework and compare ABQ with learned scale quantization (LSQ) at fixed bitwidths. As shown in Table~\ref{Ablation_ABQ}, $\lambda_b=0.0012$ yields the best performance, improving BD-PSNR by 1.68 dB and reducing BD-rate by 43.44\% over GSImage. In contrast, 6-bit LSQ shows notable redundancy. Fig.~\ref{Bitwidth_Map} illustrates that ABQ assigns higher bitwidths to complex regions and lower bitwidths to simpler areas, effectively preserving fine details while simultaneously reducing redundancy.

\begin{table}[!t]
\caption{Compression performance of different position quantization bitwidths compared with 16-bit. \label{Position_Bit_Ablation}}
\centering
\begin{tabular}{cccc}
\toprule
Dataset & Bitwidth & BD-PSNR (dB) & BD-rate (\%) \\
\midrule
\multirow{3}*{Kodak} &
14-bit & 0.1942 &  -6.4116 \\
~ & 12-bit & 0.3378 &  -10.9960 \\
~ & 10-bit &  -0.3404 & 8.2004 \\
\midrule
\multirow{3}*{DIV2K $\times 2$} &
14-bit & 0.2189 &  -6.5303 \\
~ & 12-bit & 0.3622 &  -10.7302 \\
~ & 10-bit &  -0.6969 & 16.8958 \\
\bottomrule
\end{tabular}%
\end{table}

\textbf{Position Quantization Precision.} Table~\ref{Position_Bit_Ablation} compares rate–distortion performance for different bitwidths of the position vectors. Reducing precision from 16-bit to 12-bit improves compression efficiency, achieving an 11\% BD-rate reduction and a 0.34 dB BD-PSNR gain on the Kodak dataset, and a 10.7\% BD-rate reduction and a 0.36 dB BD-PSNR gain on the DIV2K $\times 2$ dataset. This indicates that the 16-bit representations contain noticeable redundancy, while 12-bit quantization preserves essential spatial information and improves the overall RD trade-off across different datasets.



\section{Conclusion}
In this paper, we propose a structure-guided allocation principle for 2DGS representation to enhance both representational capacity and compression performance. First, we introduce a structure-guided initialization that leverages spatial structural priors to generate localized and semantically meaningful 2D Gaussian placements. Second, we adopt adaptive bitwidth quantization for covariance parameters, assigning higher precision to small-scale Gaussians in structurally complex regions, enabling RD-aware optimization while preserving critical details. Finally, we propose a geometry-consistent regularization that aligns 2D Gaussians with local gradient orientations to emphasize complex regions. Experimental results show that our method improves both representational fidelity and RD performance while maintaining real-time decoding speed.

\printbibliography

\vfill

\end{document}